%% file: selfsupervised_generalization.tex
\documentclass[10pt,twocolumn,letterpaper]{article}

\usepackage{cvpr}
\usepackage{times}
\usepackage{epsfig}
\usepackage{graphicx}
\usepackage{amsmath}
\usepackage{amssymb}
\usepackage{multicol}
\usepackage{multirow}

\usepackage[pagebackref=true,breaklinks=true,letterpaper=true,colorlinks,bookmarks=false]{hyperref}

\cvprfinalcopy % *** Uncomment this line for the final submission

\def\cvprPaperID{10221} % *** Enter the CVPR Paper ID here

\ifcvprfinal\fi
\begin{document}

%%%%%%%%% TITLE
\title{Towards Shape Biased Unsupervised Representation Learning\\for Domain Generalization}

\author{Nader Asadi\textsuperscript{\rm 1} \quad Amir M. Sarfi\textsuperscript{\rm 1} \quad Mehrdad Hosseinzadeh\textsuperscript{\rm 2} \quad Zahra Karimpour\textsuperscript{\rm 1} \quad Mahdi Eftekhari\textsuperscript{\rm 1} \\
	\textsuperscript{\rm 1} \normalsize Department of Computer Engineering, Shahid Bahonar University of Kerman, Iran\\
	\textsuperscript{\rm 2} \normalsize University of Manitoba\\
	{\tt\small \{naderasadi, m.eftekhari\}@eng.uk.ac.ir, mehrdad@cs.umanitoba.ca, \{a.m.sarfi, karimpour.zhr\}@gmail.com}
%	\and
%	Mehrdad Hosseinzade \\
%	University of Manitoba\\
%	{\tt\small mehrdad@cs.umanitoba.ca}
%	\and
%	Mahdi Eftekhari \\
%	Shahid Bahonar University\\
%	{\tt\small m.eftekhari@uk.ac.ir}
}

\maketitle
%\thispagestyle{empty}

%%%%%%%%% ABSTRACT
\input{abstractv1NA.tex}
%%%%%%%%% BODY TEXT
\input{introv1NA.tex}
\input{relatedv0.tex}
\input{methodv1MH.tex}
\section{Experiments}
In this section, we elaborate comprehensive experiments to evaluate the performance of our framework against state-of-the-art methods on domain generalization. We also present an ablation study to analyze the impact of each parameter of our framework on the exploration vs. exploitation trade-off.

\begin{table}[t!]
	\begin{center}
		\tabcolsep=0.15cm
		\begin{tabular}{l|c c c c|c}
			\hline
			\hline
			Methods&art\_paint&cartoon&sketches&photo&Avg.\\
			\hline
			\multicolumn{6}{c}{\textbf{Alexnet}}\\
			\hline
			TF\cite{li2017deeper}&62.86&66.97&57.51&89.50&69.21\\
			DeepC\cite{li2018deep}&62.30&69.58&64.45&80.72&69.26\\
			CIDDG\cite{li2018deep}&62.70&69.73&64.45&78.65&68.88\\
			MLDG\cite{li2018learning}&66.23&66.88&58.96&88.00&70.01\\
			D-SAM\cite{d2018domain}&63.87&70.70&64.66&85.55&71.20\\
			JiGen\cite{carlucci2019domain}&67.63&71.71&65.18&89.00&73.38\\
			Ours&\textbf{76.61}&\textbf{76.28}&\textbf{70.78}&\textbf{92.93}&\textbf{79.15}\\
			\hline
			\multicolumn{6}{c}{\textbf{Resnet-18}}\\
			\hline
			D-SAM\cite{d2018domain}&77.33&72.43&77.83&95.30&80.72\\
			JiGen\cite{carlucci2019domain}&79.42&75.25&71.35&96.03&80.51\\
			Ours&\textbf{83.01}&\textbf{79.39}&\textbf{78.62}&\textbf{96.83}&\textbf{84.46}\\
			\hline\hline
		\end{tabular}
	\end{center}
	\caption{Multi-source domain generalization results(\%) on PACS dataset\cite{li2017deeper} and comparison with state-of-the-art methods. Each column title represents the name of the target domain.}
	\label{table1}
\end{table}

\begin{table*}[t!]
	\begin{center}
		\def\arraystretch{1}
		\begin{tabular}{l|c c c c c c c c c c c c|c}
			\hline
			\hline
			\multirow{2}{*}{Methods}&Ar&Ar&Ar&Cl&Cl&Cl&Pr&Pr&Pr&Rw&Rw&Rw&\multirow{2}{*}{Avg.}\\
			&Cl&Pr&Rw&Ar&Pr&Rw&Ar&Cl&Rw&Ar&Cl&Pr&\\
			\hline
			ResNet-50&34.9&50.0&58.0&37.4&41.9&46.2&38.5&31.2&60.4&53.9&41.2&59.9&46.1\\
			DAN\cite{long2015learning}&43.6&57.0&67.9&45.8&56.5&60.4&44.0&43.6&67.7&63.1&51.5&74.3&56.3\\
			DANN\cite{ganin2016domain}&45.6&59.3&70.1&47.0&58.5&60.9&46.1&43.7&68.5&63.2&51.8&76.8&57.6\\
			JAN\cite{long2017deep}&45.9&61.2&6.9&50.4&59.7&61.0&45.8&43.4&70.3&63.9&52.4&76.8&58.3\\
			CDAN-RM\cite{long2018conditional}&49.2&64.8&72.9&53.8&63.9&62.9&49.8&48.8&71.5&65.8&56.4&79.2&61.6\\
			CDAN-M\cite{long2018conditional}&50.6&65.9&73.4&55.7&62.7&64.2&51.8&49.1&74.5&68.2&56.9&80.7&62.8\\
			SE\cite{french2017self}&48.8&61.8&72.8&54.1&63.2&65.1&50.6&49.2&72.3&66.1&55.9&78.7&61.5\\
			DWT\cite{roy2019unsupervised}&50.8&72.0&75.8&58.9&65.6&60.2&57.2&49.5&78.3&70.1&55.3&78.2&64.3\\
			DWT-MEC\cite{roy2019unsupervised}&54.7&72.3&77.2&56.9&68.5&69.8&54.8&47.9&78.1&68.6&54.9&81.2&65.4\\
			Jigen\cite{carlucci2019domain}&48.6&66.6&77.4&53.9&60.1&64.9&51.7&47.7&78.1&70.7&51.4&79.3&62.5\\
			Ours&\textbf{56.6}&\textbf{73.5}&\textbf{80.7}&\textbf{61.7}&\textbf{70.2}&\textbf{74.38}&\textbf{60.1}&\textbf{52.7}&\textbf{81.1}&\textbf{73.2}&\textbf{59.7}&\textbf{81.9}&\textbf{68.8}\\
			\hline\hline
		\end{tabular}
	\end{center}
	\caption{Single-source domain generalization results(\%) on Office-Home dataset\cite{venkateswara2017deep} with Resnet-50 as base network and comparison with state-of-the-art methods. Top row of each column title indicates the source domain and the bottom row represents the target domain.}
	\label{table2}
\end{table*}

\subsection{Setup}
\noindent
\textbf{Datasets}\quad 
We consider the following datasets for evaluation of our framework:

\textbf{PACS}. The PACS dataset \cite{li2017deeper} comprises four distinct domains, each corresponding to seven categories. The domains are: \textit{Photo}, \textit{Art}, \textit{Cartoon}, and \textit{Sketch}. We followed the protocol represented by \cite{carlucci2019domain} and trained our framework over multiple sources and evaluated on the target domain.

\textbf{VLCS}. The VLCS dataset \cite{torralba2011unbiased} contains five object categories shared between PASCAL VOC 2007, LabelMe, Caltech, and Sun datasets. Again, we followed the evaluation protocol used in \cite{carlucci2019domain} for multi-source generalization.

\textbf{Office-Home}. Office-Home \cite{venkateswara2017deep} is a four domain dataset, each corresponding to 65 different categories. Since there are 15,500 images in this dataset, it represents a large scale benchmark for domain adaptation problem. Using this dataset, we evaluate both multi-source and single-source generalization of our method. Correspondingly, we follow \cite{carlucci2019domain} and \cite{roy2019unsupervised} for each section. 

\textbf{MNIST $\leftrightarrow$ SVHN }. MNIST dataset \cite{lecun1998gradient} contains grayscale digits(28$\times$28) ranging from 0 to 9. On the other hand, SVHN \cite{netzer2011reading} is a color dataset with 32$\times$32 images.  However, in order to have a comparison, we scale MNIST images to 32$\times$32 treated as RGB. We followed the protocol used by \cite{roy2019unsupervised}.

\vspace{2mm}
\noindent
\textbf{Implementation Details}\quad 
To fairly demonstrate the effectiveness of our framework in improving shape bias property of unsupervised representation learning, we adopt the same base networks proposed in \cite{carlucci2019domain}. In all of our experiments, we train the networks using SGD optimizer with a batch-size of 128 images, an initial learning rate of $0.001$, weight decay of $5 \times 10^{-5}$, and momentum value of 0.9. For each experiment, the value of hyper-parameters related to either the jigsaw task or exploration vs. exploitation trade-off is stated accordingly.
For domain diversification model, we used \textit{The Behance Artistic Media Datasets(BAM)} \cite{Wilber_2017_ICCV} due to the availability of a wide variety of domains. Another salient advantage of \textit{BAM} dataset is the uniform distribution of samples across different domains; which is of significant importance to our approach since it enables the framework to create a dynamic uniform environment across arbitrary domains.

\begin{table}[t!]
	\begin{center}
		\tabcolsep=0.15cm
		\begin{tabular}{l|c c c c|c}
			\hline
			\hline
			Methods&Caltech&Labelme&Pascal&Sun&Avg.\\
			\hline
			DeepC\cite{li2018deep}&87.47&62.06&63.97&61.51&68.89\\
			CIDDG\cite{li2018deep}&88.83&63.06&64.38&62.10&69.59\\
			CCSA\cite{motiian2017unified}&92.30&62.10&67.10&59.10&70.15\\
			SLRC\cite{ding2017deep}&92.76&62.34&65.25&63.54&70.15\\
			TF\cite{li2017deeper}&93.63&63.49&69.99&61.32&72.11\\
			MMD-AAE\cite{li2018domain}&94.40&62.60&67.70&64.40&72.28\\
			D-SAM\cite{d2018domain}&91.75&56.95&58.59&60.84&67.03\\
			JiGen\cite{carlucci2019domain}&96.93&60.90&70.62&64.30&73.19\\
			Ours&\textbf{98.11}&\textbf{63.61}&\textbf{74.33}&\textbf{67.11}&\textbf{75.79}\\
			\hline\hline
		\end{tabular}
	\end{center}
	\caption{Multi-source domain generalization results(\%) on VLCS dataset\cite{torralba2011unbiased} with Resnet-18 as base network and comparison with state-of-the-art methods. Each column title represents the name of target domain.}
	\label{table3}
\end{table}

\begin{table}[t!]
	\begin{center}
		\tabcolsep=0.15cm
		\begin{tabular}{l|c c}
			\hline
			\hline
			\multirow{2}{*}{Methods}&SVHN&MNIST\\
			&MNIST&SVHN\\
			\hline
			DANN\cite{ganin2016domain}&73.9&35.7\\
			ADDA\cite{tzeng2017adversarial}&76.0&-\\
			DRCN\cite{ghifary2016deep}&82.0&40.1\\
			ATT\cite{saito2017asymmetric}&86.2&52.8\\
			ADA\cite{haeusser2017associative}&97.6&-\\
			AutoDIAL\cite{cariucci2017autodial}&89.12&10.78\\
			SBADA-GAN\cite{russo2018source}&76.1&\textbf{61.1}\\
			GAM\cite{huang2018domain}&74.6&-\\
			MEGA\cite{morerio2017minimal}&95.2&-\\
			DWT\cite{roy2019unsupervised}&\textbf{97.7}&28.9\\
			Jigen\cite{carlucci2019domain}&57.6&33.8\\
			Ours&71.7&53.7\\
			\hline\hline
		\end{tabular}
	\end{center}
	\caption{Single-source domain generalization results(\%) on MNIST and SVHN datasets with LeNet as base network. Top row of each column title indicates the source domain and the bottom row represents the target domain.}
	\label{table4}
\end{table}

\subsection{Results}\label{results}
In this section, we present an extensive experimental analysis of our framework. First, we compare our approach with state-of-the-art methods on the aforementioned benchmarks.
Second, we conduct an extensive ablation study to demonstrate the impact of our framework on leaning a robust feature representation.

\vspace{2mm}
\noindent
\textbf{Multi-Source Domain Generalization}\quad
We start our experiments by comparing our framework with state-of-the-art methods on PACS dataset\cite{li2017deeper}. 
The convolutional architecture of our approach is the same as the main structure of Alexnet or Resnet.
As discussed in section \ref{sec_framework}, our framework creates a dynamic environment for the model to explore arbitrary domains autonomously by shifting each pile of jigsaw puzzle to an arbitrary domain of BAM dataset. Table \ref{table1} demonstrates the performance of our method as well as a comparison with state-of-the-art methods on this dataset.
In this experiment, the $\rho$ parameter value, exploration rate, is of 0.5(50\%) and the $\gamma$ parameter, domain shift magnitude, is chosen between 0.5 and 0.75 randomly.
As one can see, our framework enforces the self-supervision signals to focus on the general shape of objects rather than local textures.
The significance of enhancement in learned feature representation is salient for challenging domains such as \textit{sketches} or \textit{art\_paint}.

Table \ref{table3} demonstrates the performance of our framework on VLCS\cite{torralba2011unbiased} dataset using backbone architecture of Resnet-18, same as \cite{carlucci2019domain}. We used the same setting as the previous experiment. From the table, it can be observed that our method outperforms the existing methods on VLCS \cite{torralba2011unbiased} dataset with a considerable margin.

\begin{figure*}[t!]
	\begin{center}
		\includegraphics[height=3.75cm]{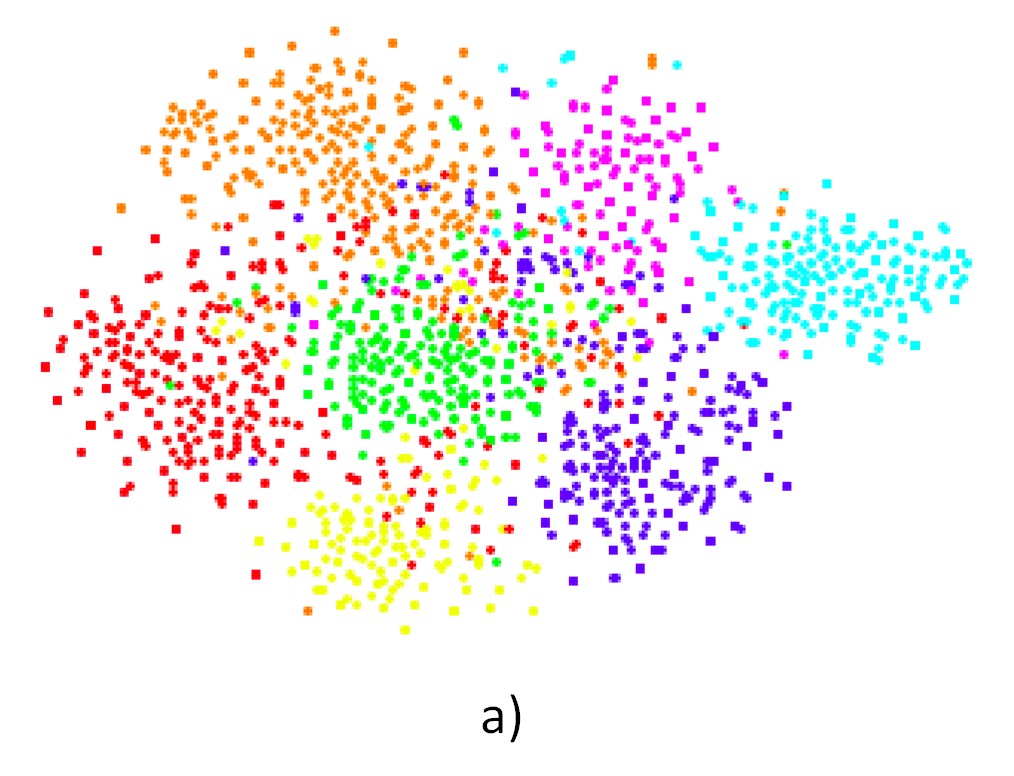}
		\hspace{3mm}
		\includegraphics[height=3.75cm]{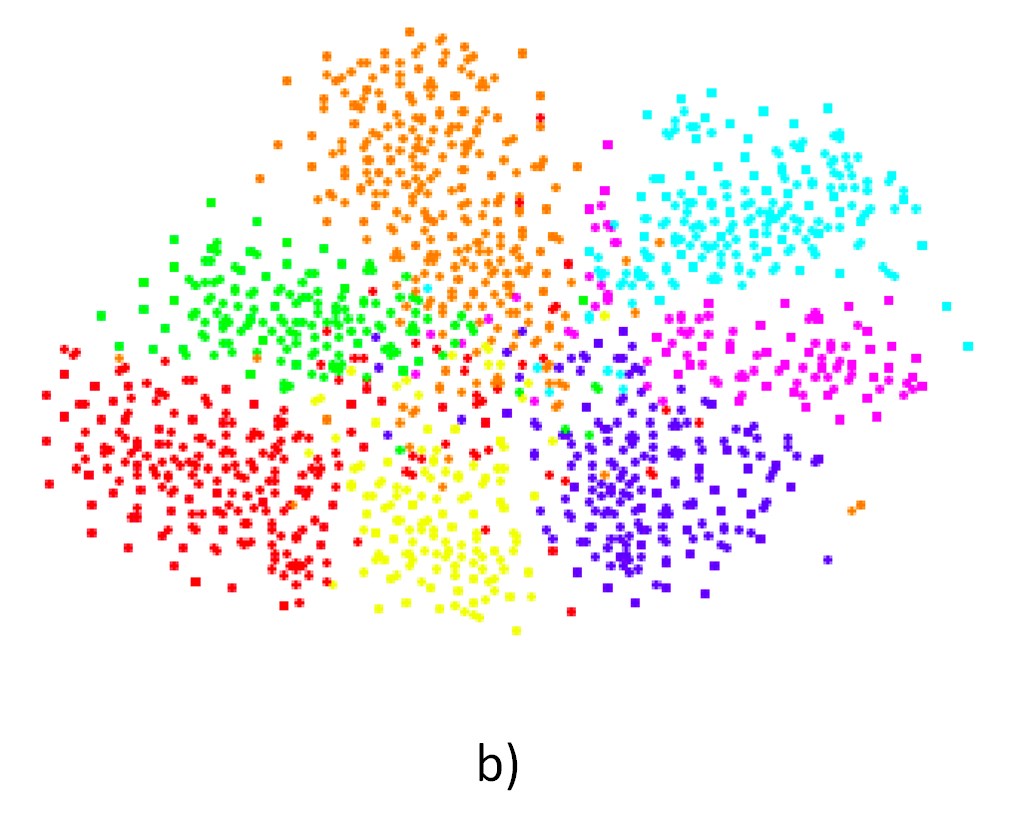}
		\hspace{3mm}
		\includegraphics[height=3.75cm]{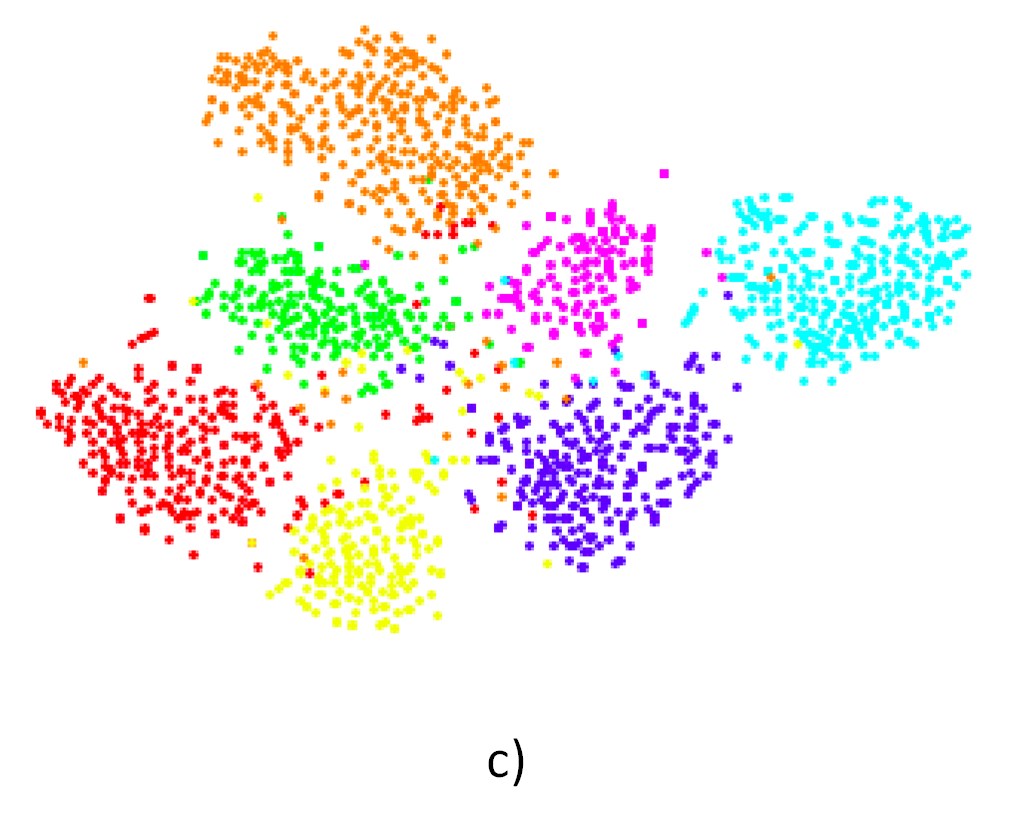}
	\end{center}
	\caption{t-SNE visualizations of the feature representations for \textit{art\_painting} as target domain of PACS dataset. Obviously, (a) supervised learning barely captures any robust classification patterns. (b) Though self-supervised learning(JiGen\cite{carlucci2019domain}) improves learned feature representation, it is not sufficiently robust against domain shifts. However, (c) our method enforces the model to learn shape biased and domain-invariant representations.}
	\label{figure4}
\end{figure*}

\vspace{2mm}
\noindent
\textbf{Single-Source Domain Generalization}\quad
We compare the performance of our framework with several methods on single-source domain generalization. We used \cite{roy2019unsupervised} as reference for the performance of other methods. However, for Jigen\cite{carlucci2019domain}, we used their public source code. Table \ref{table2} demonstrates the prominence of our method on Office-Home\cite{venkateswara2017deep} dataset. As one can see, our method outperforms other methods in all domains by a considerable margin. Also, table \ref{table4} presents a comparison on MNIST $\leftrightarrow$ SVHN generalization problem.

\begin{figure}[t!]
	\begin{center}
		\includegraphics[width=5.5cm]{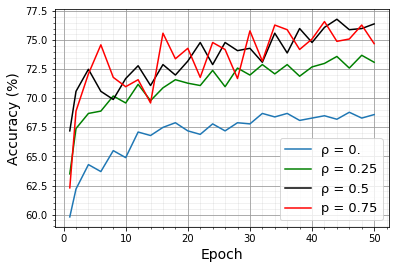}
	\end{center}
	\caption{Analysis of the behavior of our framework with different $\rho$ values on PACS dataset \cite{li2017deeper} with \textit{art\_painting} as target domain. The $\gamma$ value is equal to $0.75$ and Alexnet is used as the backbone network.}
	\label{figure5}
\end{figure}

\subsection{Ablation Study}

In this section, we perform thorough ablation experiments to investigate the effect of different modules and parameters in our framework. These experiments demonstrate the contributions of different modules and provide more insight into our approach.
All experiments of this section are conducted on PACS dataset \cite{li2017deeper} with art\_painting as the target domain and Alexnet architecture is used as the backbone network.

\vspace{2mm}
\noindent
\textbf{Activation Visualization}\quad 
In this experiment, we visualize feature representation at the final layer of the convolutional network(Alexnet). As can be seen in figure \ref{figure4}, deep network alone barely captures any useful classification patterns relevant to the target domain. Though vanilla self-supervision signals can help improve learned feature representation, they are biased to superficial statistics of data. Since our framework tackles the unfavorable biases of self-supervised methods toward textures and local features, it helps the model learn shape biased representations which are robust across domains.

\begin{figure}[t!]
	\begin{center}
		\includegraphics[width=5.5cm]{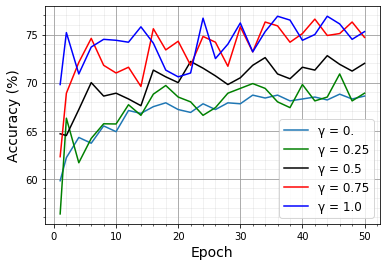}
	\end{center}
	\caption{Analysis of the behavior of our framework with different $\gamma$ values on PACS dataset \cite{li2017deeper} with \textit{art\_painting} as target domain. The $\rho$ value is equal to $0.75$ and Alexnet is used as the backbone network.}
	\label{figur6}
\end{figure}

\begin{figure*}[t!]
	\begin{center}
		\includegraphics[width=17cm]{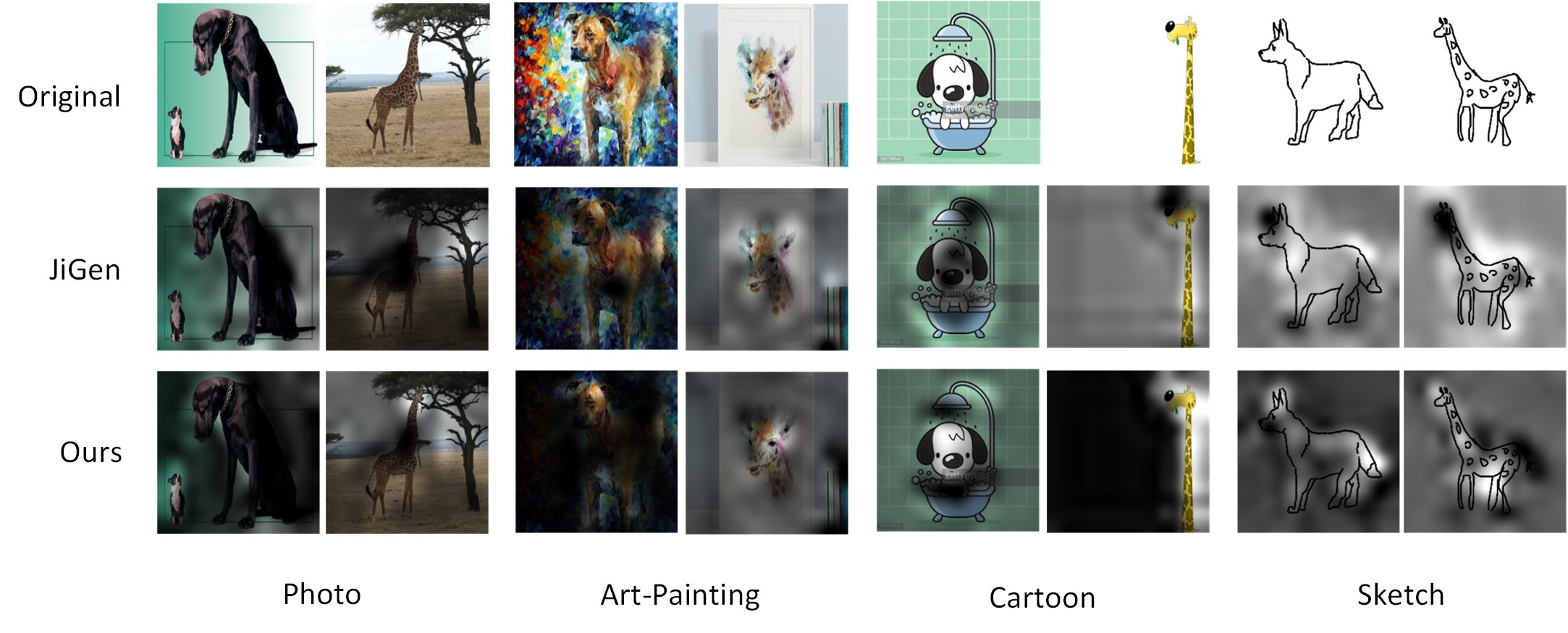}
	\end{center}
	\caption{Attention maps generated by our framework(third row) and JiGen\cite{carlucci2019domain}(second row) on PACS dataset\cite{li2017deeper} using Alexnet as backbone network. As one can observe, our framework enforces the model to focus on the general shape of objects. As a result, the value of attention maps on the background and context of the image has diminished.}%pls note that as it gets deeper, samples get further from their correct class
	\label{figure7}
\end{figure*}

\vspace{2mm}
\noindent
\textbf{Exploration vs Exploitation Analysis}\quad
In this experiment, we perform ablation study for investigating the effect of exploration vs. exploitation trade-off parameters, namely $\rho$ and $\gamma$. The $\rho$ parameter defines exploration rate across arbitrary domains. With a high value of $\rho$, the agent(model) is mostly exploring random domains rather than exploiting static source domains. Figure \ref{figure5} illustrates the impact of $\rho$ parameter on the learning process of the network. 
As one can observe, with high values of $\rho$ (0.75 and above), the accuracy on test-set fluctuates considerably, which hinders the convergence process of the network. Conversely, low values of $\rho$ (0.25 and below) do not allow the agent to sufficiently explore new domains. As a result, the shape bias property of the model is not improved enough, and the learned feature representation is not robust against domain shifts.

The second and equally significant parameter is $\gamma$, which defines the magnitude of domain translations during the exploration phase. Same as $\rho$, high values of $\gamma$ results in substantial fluctuations in test-loss during the training process of the network. On the other hand, low values of $\gamma$ do not allow the model to shift toward arbitrary domains sufficiently. Figure \ref{figur6} clearly demonstrates the effect of $\gamma$ parameter.
For all experiments in section \ref{results}, we used $\rho$ value of $0.5$ and $\gamma$ was chosen randomly between $0.75$ and $1.0$.
Another satisfactory approach is to initialize $\rho$ and $\gamma$ parameters with high values and gradually diminish them for better convergence.

\vspace{2mm}
\noindent
\textbf{Attention Visualization}\quad
In this experiment, we visualize some attention maps for the network trained by our method and JiGen\cite{carlucci2019domain} to provide more insight into our approach.
These attention maps are computed based on the magnitude of
activations at each spatial cell of a convolutional layer and essentially reflect where the network
puts most of its focus in order to classify an input image.
Figure \ref{figure7} provides a comparison between attention maps generated by our framework and JiGen\cite{carlucci2019domain}.
As discussed in section \ref{sec_framework}, our framework enforces the network to focus on the general shape of objects rather than superficial statistics of context e.g., textures.
Figure \ref{figure7} clearly demonstrates the prominence of shape bias and domain-invariance properties of our method.

\section{Conclusion and Future Works}
In this paper, we studied unfavorable biases of unsupervised representation learning and introduced a learning paradigm to alleviate this problem. Our framework achieves this goal by integrating domain diversification and self-supervision signals. 
Domain diversification module creates a dynamic environment across arbitrary domains for the agent to explore autonomously.
Also, our framework provides an exploration vs. exploitation trade-off setting. Finally, we formulized our framework for domain generalization problem and conducted extensive experiments and analysis to demonstrate the effectiveness of our method.

There are many opportunities for contributing to this research thread. The key idea behind this paper was to increase the integration between agent and environment by creating a dynamic environment across arbitrary domains so that the agent can explore this environment autonomously rather than just exploiting a static source dataset. However, the exploration phase can be improved drastically using Meta-Learning or Reinforcement Learning methods to robustify learned feature representations against domain shifts.

{\small
\bibliographystyle{ieee_fullname}
\bibliography{egbib}
}

\end{document}

%% file: abstractv1NA.tex
\begin{abstract}
   Shape bias plays an important role in self-supervised learning paradigm. The ultimate goal in self-supervised learning is to capture a representation that is based as much as possible on the semantic of objects (\textit{i.e.} shape bias) and not on individual objects' peripheral features. This is inline with how human learns in general; our brain unconsciously focuses on the general shape of objects rather than superficial statistics of context. On the other hand, unsupervised representation learning allows discovering label-invariant features which helps generalization of the model. Inspired by these observations, we propose a learning framework to improve the learning performance of self-supervised methods by further hitching their learning process to shape bias. Using distinct modules, our method learns semantic and shape biased representations by integrating domain diversification and jigsaw puzzles. The first module enables the model to create a dynamic environment across arbitrary domains and provides a domain exploration vs. exploitation trade-off, while the second module allows it to explore this environment autonomously. The proposed framework is universally adaptable since it does not require prior knowledge of the domain of interest. We empirically evaluate the performance of our framework through extensive experiments on several domain generalization datasets, namely, PACS, Office-Home, VLCS, and Digits. Results show that the proposed method outperforms the other state-of-the-arts on most of the datasets.
\end{abstract}

%% file: introv1NA.tex
\section{Introduction}
Convolutional Neural Networks have achieved significant success in a wide variety of visual recognition tasks, demonstrating an excellent ability to learn the spatial structure of image data.
Despite the promising results, the performance of these models diminishes considerably when applied to environments different than training. This fragile phenomenon is known as \textit{domain shift}.
% and is well-studied in the literature
This is due to the significant dependency of these models on large-scale annotated datasets. This issue imposes severe issues toward practical applications of deep neural networks and their generalization to domains beyond their own training one \cite{torralba2011unbiased}.

%\begin{figure}[h!]
%	\begin{center}
%		\includegraphics[width=6.5cm]{assets/jigsaw1.jpg}
%	\end{center}
%	\caption{Superficial statistics of image, e.g., textures, can help solve jigsaw puzzles which leads to unfavorable biases of the network. We propose a learning framework which creates a controllable dynamic environment across arbitrary domains and enables the agent(model) to explore this environment autonomously.}%pls note that as it gets deeper, samples get further from their correct class
%	\label{figure1}
%\end{figure}

In order to alleviate \textit{domain shift} problem, several domain adaptation methods have been proposed \cite{csurka2017domain, pan2009survey, wang2018deep}
Due to the cost of manually annotating target domain data, unsupervised domain adaptation, \textit{i.e.} adapting a trained model to a new target domain without annotating the target dataset, has been the foundation for a considerable amount of research.
Early works in this domain tackled the problem by learning to match the distribution of source and target domains \cite{sun2016deep, morerio2017minimal}.
In these methods, the goal is to reduce the difference between covariance matrices of source and target domains.
Similarly \cite{roy2019unsupervised} proposed Domain-specific Whitening Transform which computes domain-specific covariance matrices of intermediate
features by whitening the source and the target
features and projecting them into a common spherical distribution.
%[[next two paragraphs focuses too much on related work without linking them to our method; move them to sec. 2 or link them to our idea. remember intro. is where you can catch or lose your reviewer!]]\\
Another recent approach used for unsupervised domain adaptation, embeds domain-specific alignment layers, inspired by BatchNorm\cite{ioffe2015batch} layers, within the network \cite{cariucci2017autodial,mancini2018boosting}
Another category of domain translation methods focuses on the appearance translation of source domain towards the target domain.
Since many image translation methods are imperfect, these approaches have severe drawbacks especially when there is a significant gap between source and target domains \cite{bousmalis2017unsupervised, hoffman2017cycada, russo2018source}.

Due to the fact that unsupervised representation learning helps the model to learn label-invariant and high-level representation of data, they can improve generalizability as well as the robustness of the model \cite{carlucci2019domain, hendrycks2019using}. 
Furthermore, unsupervised representation learning can solve the lack of large-scale annotated benchmarks since a sheer volume of unlabeled data is publicly available. Subsequently, these methods are less susceptible to unfavorable biases of data \cite{torralba2011unbiased}.
Inspired by what mentioned above, \cite{carlucci2019domain} proposed a method capable of learning spatial co-location of image parts simultaneously with supervised learning. Moreover, they show that self-supervision can have a considerable impact on the generalizability of the model.
Although self-supervised methods learn general feature embeddings, one major question is how much is the learned  representation by these methods biased towards superficial statistics of image \textit{e.g.}, textures and local features?

%[[this paragraph about human is wandering around like a youtube ad. move it to more appropriate place]]\\
%[[too much related work again. try to pitch your own method]]\\

It is proved that humans' biological visual system has a considerable robust performance against domain shifts. It is known that humans display shape bias when classifying new objects and that is the reason behind the robustness of human's visual system \cite{ritter2017cognitive, geirhos2018generalisation}.
In recent years, shape-bias property and unfavorable bias of CNNs towards textures and local features has been studied extensively \cite{hosseini2018assessing, geirhos2018imagenet, brendel2019approximating}.
Our model is partly inspired by some work of Geirhos et al. \cite{geirhos2018imagenet}. Geirhos et al. \cite{geirhos2018imagenet} proves that CNNs are highly biased towards learning superficial characteristics e.g., textures rather than general shape of objects which is in clear contrast with the human visual system.
In this paper, we use a similar idea and develop a CNN architecture to improve the generalization of Self-supervised methods against arbitrary domain shifts.

Furthermore, we investigate the learned feature representation of self-supervised methods. We state that, though unsupervised representation learning is capable of learning a label-invariant and high-level representation of the image, it is severely biased to unfavorable and superficial statistics of the data. We use jigsaw puzzle \cite{noroozi2016unsupervised} pretext task as the baseline of our method and analysis. We prove that context-based self-supervised methods are significantly biased to the texture and local features of the image.

To remedy it, a novel shape-biased framework for self-supervised methods is proposed as a means to learn more generalized representation and reduce domain shift problem. A texture diversification method is used in order to enforce the model to solve jigsaw task based on the general shape of each object rather than local textures and superficial features. Thereupon, we propose a self-supervised learning system based on jigsaw task which enables the model to explore arbitrary domains during the training process.
Furthermore, an extensive analysis is conducted on domain exploration vs. exploitation trade-off, and the behavior of the model is discussed in each setting. 

Since self-supervised methods learn an invariant representation of the data, improving shape bias property, and unbiasing them from superficial statistics, enhances the generalizability of these models to a significant extent. Finally, in order to prove the effectiveness of our method, we conduct extensive experiments on domain generalization benchmarks and illustrate that our method outperforms state-of-the-art works on domain generalization.
To summarize the contribution of this paper is manifold:
\begin{itemize}
	\item We analyze the dependency of unsupervised representation learning methods on textures as well as local features of image. 
	\item We structurize an unsupervised representation learning paradigm by integrating domain diversification and jigsaw puzzles pretext task.
	\item Through extensive experiments, we show that our method outperforms state-of-the-art methods on domain generalization. Furthermore, the robustness of our method against distortions is studied.
\end{itemize}

\begin{figure*}[h!]
	\begin{center}
		\includegraphics[width=16cm]{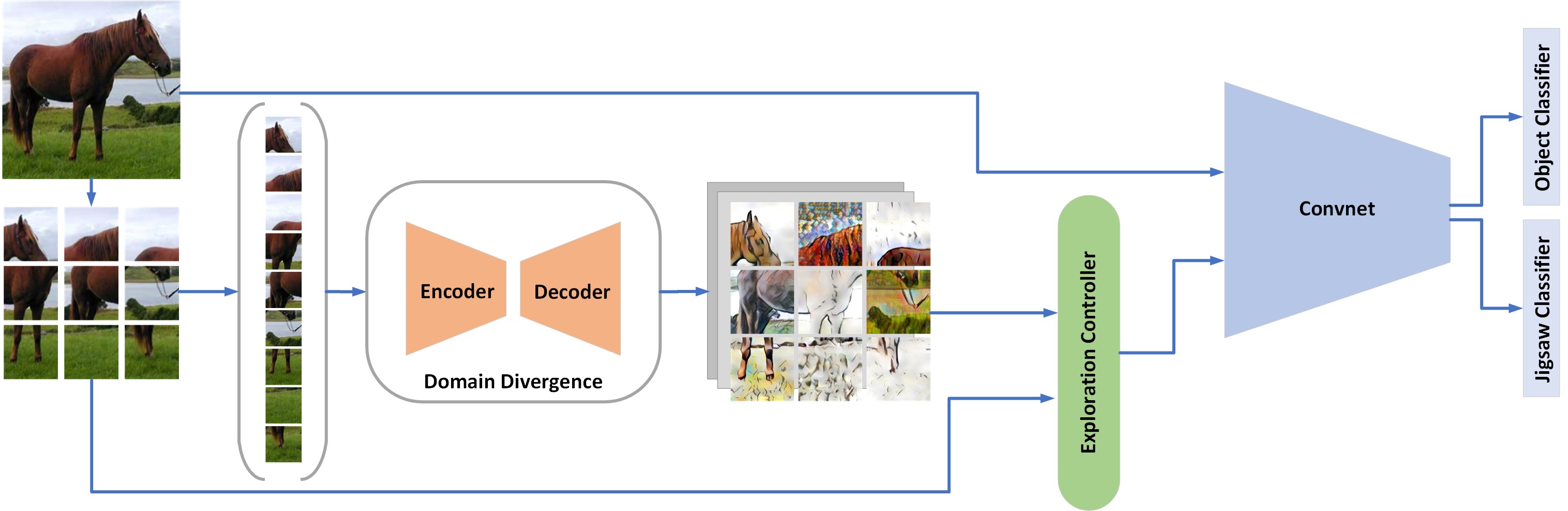}
	\end{center}
	\caption{Illustration of our proposed framework. Our framework enables the model(Convnet) to learn from both ordered and shuffled images. Our framework consists of Domain Diversification followed by Exploration Controller modules which create a dynamic environment across arbitrary domains. Consequently, the model can explore this environment autonomously, which improves the shape bias property of the network to a significant extent.}%pls note that as it gets deeper, samples get further from their correct class
	\label{figure2}
\end{figure*}

%% file: relatedv0.tex
\section{Related Works}
As discussed in the previous section, in this work, we study the generalization of self-supervised methods across arbitrary domains. Correspondingly, a model based on Jigsaw pretext task is proposed as a domain generalization method. Our work encompasses DomainDomain Adaptation and Unsupervised Representation Learning problems. We briefly discuss each problem in this section.
\subsection{Unsupervised Domain Adaptation}
Domain adaptation, the challenge of distilling the most general and transferable knowledge from a limited source, has been studied intensively for both shallow and deep neural networks. However, in this section, our primary focus is on deep domain adaptation methods due to the significant success of these models\cite{jing2019self}.

Generally, these methods lay into generative-based methods\cite{zhang2016colorful, pathak2016context, zhu2017unpaired}, free-semantic based methods\cite{ren2018cross}, and context-based methods\cite{noroozi2016unsupervised, gidaris2018unsupervised, kim2018learning}.
One category of research tackles the domain adaptation problem using a set of unannotated target data to guide training on the source domain \cite{csurka2017domain}. However, real-world applications of domain generalization target domain data are not available, which is considered as a drawback of the aforementioned approaches. Some works use the first-order\cite{long2015learning, venkateswara2017deep} and second-order statistics\cite{sun2016deep, morerio2017minimal} to diminish the domain shift problem.
Some other works try to alleviate domain shift problem by introducing domain alignment layers\cite{mancini2018boosting, cariucci2017autodial}, inspired by BachNorm layer\cite{ioffe2015batch}.

Another category of feature level approaches work on Maximum Mean Discrepancy minimization\cite{long2015learning, sun2016deep, cariucci2017autodial}. Using Generative Adversarial Networks(GANs), source-to-target transformation methods have shown promising results \cite{zhu2017unpaired, kim2017learning, zhu2017unpaired}. However, since translation models are imperfect, GAN-based methods have significant drawbacks in case of complex datasets.
Several unsupervised domain adaptation methods used \textit{entropy minimization} \cite{grandvalet2005semi}, exploiting the high-confidence predictions of unlabeled samples as pseudo-labels, due to the effectiveness of this method \cite{long2017deep, saito2017asymmetric, russo2018source}.

\subsection{Unsupervised Representation Learning}
Self-supervised learning is a framework in which the model is explicitly trained with automatically generated labels from \textit{pretext task}, in an effort to learn useful representations for \textit{downstream task}. In this section, we focus on self-supervised methods for image representation learning.

Patch-based unsupervised representation learning is one of the major approaches which was first introduced by \cite{doersch2015unsupervised}.
On the same research line, \cite{noroozi2016unsupervised} proposed a method that predicts the permutation of \textit{jigsaw puzzles}. Some other methods are built on top of jigsaw task with an effort to improve the learned feature representation for downstream task \cite{noroozi2018boosting, kim2018learning}.
Other methods generate image-level tasks. \cite{gidaris2018unsupervised} proposed a method based on the random rotation of the image at specific angles.
\cite{caron2018deep} used clustering of images in the latent space of the network to generate pseudo-labels for training the model.
Other category of works are based on generative-based methods. Some noteworthy examples are super-resolution\cite{ledig2017photo}, image inpainting \cite{pathak2016context}, image colorization\cite{zhang2016colorful}, and generative adversarial networks\cite{zhu2017unpaired}.

\cite{carlucci2019domain} proposed a method with respect to previous literature by investigating the importance of jointly exploiting supervised
and unsupervised inherent signals from the images for domain generalization. However, we study the unfavorable biases of learned feature representation of self-supervised methods. We propose a method in an effort of focusing on domain-agnostic signals and improving the shape bias property of these models. 

%% file: methodv1MH.tex
\section{Methodology}
It is known that the human visual system generally displays shape bias.
On the other hand, humans seem to rely crucially on learning autonomously. As a matter of fact, the human visual system adaptability relies on both of the aforementioned approaches \cite{ritter2017cognitive, geirhos2018generalisation}.
This is particularly effective, since unsupervised representation learning methods are label-invariant, they can help the model to discover invariances and regularities that help to generalize \cite{carlucci2019domain}.
However, the learned feature representation in these methods is highly dependent on the pretext task e.g., jigsaw puzzles which might cause unfavorable biases in learned feature representation. In order to solve this task, the model can also use superficial statistics which severely harm shape bias property as well as the generalizability of the model.

We propose a learning paradigm to improve the shape bias property of unsupervised representation learning by alleviating unfavorable biases to superficial statistics of the image.
Our framework is based on domain diversification and self-supervision together, which allows the model to be free from a single static domain. In other words, the agent(model) is able to autonomously explore a dynamic environment across arbitrary domains in an effort of learning domain invariant and shape biased characteristics of the problem.
Also, we demonstrate that exploration vs. exploitation trade-off can help the model learn useful and unbiased representations.
Our method has two major modules, Domain Diversification followed by exploration vs. exploitation controller and unsupervised representation learning by solving jigsaw puzzles. In the following, we describe each module in detail, and finally, a domain invariant learning framework is introduced. Figure \ref{figure2} demonstrates conceptual description of our method.

\subsection{Jigsaw Puzzles for Domain Generalization}
Previous work \cite{carlucci2019domain} has shown the efficiency of solving jigsaw puzzles on learning a general representation of the image. We use the same approach as the baseline of our framework.
This module consists of a convolutional feature extractor and two fully connected parts.
Consider $x$ as the input of the network, $y$ as the correct label, and $\theta_f$, $\theta_c$, $\theta_j$ as parameters of convolutional feature extractor, cross-entropy classifier, jigsaw classifier respectively.
The objective of the first fully-connected module is to minimize the cross-entropy loss between ground-truth label $y$ and predicted label by the model parameterized by $\theta_f$ and $\theta_c$. This module, as well as the feature extractor, are trained through
\begin{equation}
\underset{\theta_f, \theta_c}{argmin}
\sum\limits_{j=1}\limits^{N_i}
L_c (h(x_j^i | \theta_f, \theta_c), y_j^i)
\end{equation}

The objective of the second module is to predict permutation set in a decomposed image of $n \times n$ grid of patches being randomly shuffled. The overall number of possible permutations is $n^2!$, however, similar to \cite{carlucci2019domain}, a set of P elements is selected following the Hamming distance-based algorithm presented in \cite{noroozi2016unsupervised}. This module, as well as the feature extractor, are trained through
\begin{equation}
\underset{\theta_f, \theta_j}{argmin}
\sum\limits_{k=1}\limits^{K_i}
L_j (h(z_k^i | \theta_f, \theta_j), p_k^i)
\end{equation}
where $z_k^i$ indicates the decomposed samples and $p_k^i$ the corresponding permutation index.

\subsection{Domain Diversification}
Domain Diversification is a method which diversifies the source domain by intentionally generating distinctive domain discrepancy through these domain shifters.
Without harming general shape of the object, the domain of each jigsaw puzzle tile is shifted arbitrarily.
Consequently, superficial statistics of the image like textures, have no benefits for solving the pretext task. Thus, the model is enforced to extract features relevant to the general shape of each object. As a result, the self-supervised model will be biased to the high-level shape of objects rather than local textures and background.

Due to the imperfect performance of domain-shifters, they can cause issues when directly used for domain adaptation problems. However, our framework is utilized by domain diversification to create a dynamic environment in which the self-supervised module is enforced to learn domain-invariant representations. So, we use a modification of \textit{Adaptive Instance Normalization} method due to its high speed and acceptable performance.
%Since some of the hyper-parameters of AdaIN have a significant impact on not only exploration vs. exploitation trade-off but also the overall performance of our framework, we briefly discuss this method. For further details, we refer the reader to \cite{huang2017arbitrary}.
Consider $x^s$ as the feature representation of source data and $x^t$ as the feature representation of an arbitrary target domain. In general, style transfer methods align the channel-wise mean and variance of $x^s$ to match those of $x^t$.
\begin{equation}
	\textnormal{AdaIN}(x^s, x^t) = \sigma(x^t)(\frac{x^s - \mu(x^s)}{\sigma(x^s)}) +
	\mu(x^s)
\end{equation}\label{eq3}
as one can see, AdaIN simply scales the normalized content input with $\sigma(x^t)$, and shift it with $\mu(x^t)$. It is crucial to keep the overall shape of each tile intact. The loss function below matches the style of source image to the target image.
\begin{multline}
L_s(t, s) = \sum_{i = 1}^{N}\|\mu(\phi_i(g(t))) - \mu(\phi_i(s))\|_2 +\\ 
\sum_{i = 1}^{N}\|\sigma(\phi_i(g(t))) - \sigma(\phi_i(s))\|_2
\end{multline}
where $\phi$ denotes a layer in VGG-19 used to computer the style loss. Meanwhile, adjacent tiles should be of distant domains in order to enforce the self-supervised model learn shape biased representations.
Moreover, since the texture of each tile is changed randomly in each iteration of training, the superficial statistics of the image have no benefits for solving the jigsaw puzzle. To redeem this problem, we add an objective function to enforce the model diversify adjacent tiles without harming the shape of objects.
\begin{equation}
L_{j} = \sum_{j = 2}^{K - 1}L_s(z_k, z_{k-1}) + L_s(z_k, z_{k+1}) 
\end{equation}
where $K$ is the number of jigsaw tiles and $z_k$ indicates the decomposed samples . A pretrained VGG-19 is used to compute the loss function:
\begin{equation}
L = L_c + \lambda L_s - \tau L_{j}
\end{equation}\label{eq5}
which is a weighted combination of the content loss and style losses where $L_c$ is
\begin{equation}
L_c = \|f(g(t)) - t\|_2
\end{equation}\label{eq6}
In order to keep the overall shape of the object intact, $\lambda$ and $\tau$ parameters should be chosen smaller compared to style transfer problems.
Finally, the output of translation module is computed as:
\begin{equation}
	T({x^s}, {x^t}, \gamma) = g((1 - \gamma)f(x^s) + \gamma \textnormal{AdaIN}(f(x^s), f(x^t)))
\end{equation}\label{eq7}
where $f$ is the pretrained VGG19 encoder, $g$ is the trained decoder using Eq \ref{eq3}, and $\gamma$ allows content-style trade-off at test time.

%\vspace{2mm}
%\noindent
%\textbf{Structured Framework}\quad 
\subsection{Structured Framework}\label{sec_framework}
In this section, we structurize our learning paradigm by integrating Domain Diversification and solving jigsaw puzzles into a framework. As previously discussed, though self-supervised methods learn general and high-level representation of data, they can be highly biased to superficial statistics of the image e.g., textures and local features. The objective of our work is to create a dynamic environment across domains for the agent(model) to autonomously learn texture-invariant and shape-biased statistics of the data.
To achieve this goal, we randomly shift each jigsaw tile to an arbitrary domain. Subsequently, the superficial context and texture of the piles have no advantage for solving the pretext task; thus, the model is enforced to rely on the overall shape of each object to solve the task. The overall objective function of the framework can be written as follows:
\begin{multline}
	\underset{\theta_f, \theta_c, \theta_j}{argmin}
	\sum\limits_{i=1}\limits^{N}
	L_c (h(x_i | \theta_f, \theta_c), y_i)
	 + \\
	\sum\limits_{k=1}\limits^{K}
	\alpha L_j (h(D(z_k, \rho) | \theta_f, \theta_j), p_k)
\end{multline}
Here, $N$ is the number of labeled instances, $z_k$ indicates the recomposed samples for jigsaw task, $\alpha$ is the weight parameter for jigsaw loss, $\rho$ is the domain shift probability parameter, and $D$ stands for domain diversification model trained using Eq\ref{eq5}.

\vspace{2mm}
\noindent
\textbf{Exploration vs Exploitation Trade-off}\quad \\
Our framework has four important hyper-parameters. Two of which, namely $\alpha$ and $\beta$, are related to solving jigsaw puzzles \cite{carlucci2019domain}. The $\alpha$ parameter, as mentioned before, controls the significance of jigsaw loss on the overall objective function. The $\beta$ parameter defines the ratio between ordered and shuffled images.

Two other parameters, namely $\gamma$ and $\rho$, are highly correlated with exploration vs. exploitation trade-off.
The $\rho$ parameter indicates the probability of domain shift, in other words, exploration weight. For instance, $\rho = 0.7$ means that for each batch, 70\% of the shuffled images have arbitrarily shifted patches. On the other hand, $\gamma$ (Eq. \ref{eq5}) defines the magnitude of texture translation; in other words, the step size towards arbitrary domain at exploration time.
These two parameters enable the model to explore a dynamic environment across arbitrary domains autonomously.
In the following section, we conduct extensive analysis on the impact of these parameters on the learned feature representation of the network.